\documentclass[runningheads]{llncs}
\usepackage[T1]{fontenc}

\usepackage{graphicx}
\usepackage{amsmath}
\usepackage{amsfonts}
\usepackage{subcaption}
\usepackage{float}
\newcommand{\parsection}[1]{\textbf{#1}}

\begin{document}
\title{Continuous Normalizing Flows for Uncertainty-Aware Human Pose Estimation}
\titlerunning{CNFs for Uncertainty-Aware Human Pose Estimation}
% If the paper title is too long for the running head, you can set
% an abbreviated paper title here
%
%\author{Anonymous Submission to SCIA 2025}
\author{Shipeng Liu\orcidID{0009-0004-0703-9431} \and Ziliang Xiong\orcidID{0009-0008-8277-7476} \and  Bastian Wandt\orcidID{0009-0002-1203-1093} \and  Per-Erik Forssén\orcidID{0000-0002-5698-5983}}
%
%\authorrunning{I. M. Anonymous et al.}
\authorrunning{Shipeng Liu et al.}
% First names are abbreviated in the running head.
% If there are more than two authors, 'et al.' is used.
%
\institute{Computer Vision Laboratory, Dept of E. E., Linköping University, Sweden\\ \email{\{firstname.lastname@liu.se\}}\\ }%\email{\{ziliang.xiong, bastian.wandt, per-erik.forssen\}@liu.se}\\ }
\maketitle              % typeset the header of the contribution
\vspace{-6mm}
\begin{abstract}
Human Pose Estimation (HPE) is increasingly important for applications like virtual reality and motion analysis, yet current methods struggle with balancing accuracy, computational efficiency, and reliable uncertainty quantification (UQ). Traditional regression-based methods assume fixed distributions, which might lead to poor UQ. Heatmap-based methods effectively model the output distribution using likelihood heatmaps, however, they demand significant resources. To address this, we propose Continuous Flow Residual Estimation (CFRE), an integration of Continuous Normalizing Flows (CNFs) into regression-based models, which allows for dynamic distribution adaptation.
Through extensive experiments, we show that CFRE leads to better accuracy and uncertainty quantification with retained computational efficiency on both 2D and 3D human pose estimation tasks.

\end{abstract}
\vspace{-8mm}
\section{Introduction}

Human Pose Estimation (HPE) is a critical task in computer vision with applications spanning virtual reality, motion analysis, and intelligent surveillance. It involves identifying human keypoint locations from visual data. These keypoints correspond to anatomical landmarks such as joints (e.g.\ shoulders, elbows, and knees).
Despite steady improvements, reliable real-world performance remains a challenge due to complexities such as noise, occlusions, and diversity in posture. Addressing these issues requires not only accurate pose estimation but also effective uncertainty quantification to ensure robustness and interpretability in practical scenarios.

Uncertainty Quantification (UQ) is particularly significant in dynamic and unpredictable environments, such as safety-critical applications. Beyond predicting joint coordinates, understanding the uncertainty of the model in its outputs enables error identification, adaptation to abnormal input, and informed decisions on the need for human intervention. However, existing methods often fail to sufficiently address uncertainty estimation.

Two dominant approaches in HPE are regression-based and heatmap-based methods. Regression-based methods \cite{li_human_2021,zhou_objects_2019,9010416,10.1007/978-3-030-58607-2_31} directly predict joint coordinates by minimizing loss functions such as $\ell_1$ loss (Laplace assumption) or $\ell_2$ (Gaussian assumption). Although computationally efficient, these models are based on fixed distributional assumptions \cite{928e476715544027af08ec20936dd6ca}, which can lead to suboptimal performance and unreliable uncertainty estimates. For instance, heteroscedastic regression based on Laplace distributions has shown better results compared to Gaussian-based models \cite{li_human_2021}, indicating that real-world joint locations often deviate from Gaussian assumptions. 
% However, these methods struggle with data complexities, such as heavy-tailed distributions, which affect variance estimation \cite{gorji_robust_2020}.

Heatmap-based methods \cite{10.1007/978-3-030-01231-1_29,sun_deep_2019,8237584,10.1007/978-3-030-01264-9_17,mcnally_evopose2d_2021}, on the other hand, frame keypoint detection as a classification problem by generating likelihood heatmaps for each keypoint. Although these approaches, such as HRNet \cite{sun_deep_2019} and SimplePose \cite{10.1007/978-3-030-01231-1_29}, achieve high accuracy, they come with significant computational and memory requirements.

\begin{figure}[h]
    \centering
    \includegraphics[width=0.9\textwidth]{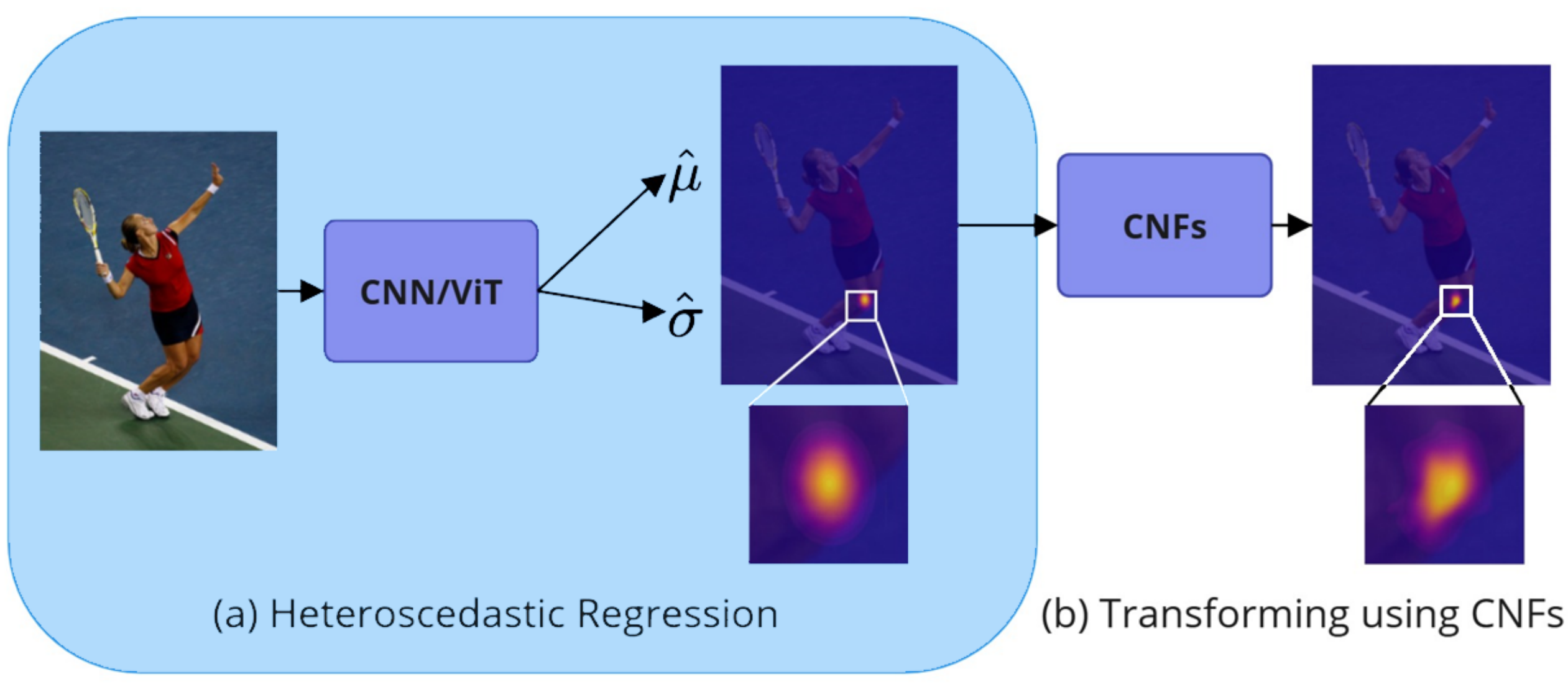}
    \caption{Overview of our proposed method. Heteroscedastic Deep Regression (Blue box) estimates the mean and scale of keypoint locations, which can be described as a simple distribution (e.g., Gaussian or Laplacian). Our CNF then transforms this into a more complex uncertainty distribution.}
    \label{fig:flow_chart}
\end{figure}

To address these limitations, we propose a novel approach that combines the strengths of regression-based methods with Continuous Normalizing Flows (CNFs) \cite{NEURIPS2018_69386f6b} shown in Figure \ref{fig:flow_chart}. 

\vspace{-6mm}

\vspace{-5mm}

A CNF uses neural ordinary differential equations to learn complex data distributions without fixed assumptions, using continuous-time transformations.
The resulting framework, Continuous Flow Residual Estimation (CFRE), integrates a prior regression network with CNFs to estimate joint locations and associated uncertainties. During training, the regression network assumes a Laplace distribution, while CNFs dynamically refine the predictions to better align with real-world data. Importantly, the framework incurs no additional computational costs during inference, making it both practical and efficient.

We validate our method on 2D and 3D HPE tasks using MSCOCO \cite{cocodataset} and Human3.6M \cite{h36m_pami}, achieving superior performance over traditional regression and heatmap-based models while ensuring calibrated confidence. Qualitative results indicate that the model captures more complex patterns, including anisotropy and sharper peaks than Gaussian distributions. %increased vertical variance. Too specific at this point
Experimental results on mAP and AUSE metrics demonstrate that improved data distribution fitting also enhances uncertainty quantification.

Our contributions include:

% 1. Human pose， uncertainty calibration
% 2. heatmap， regression，fixed distribution with commn loss
% 3. Using heteroscadetic regression, reveal misalignment, 
% 4. inspired by RLE, we study CNF, starting from maximum likelihood, to flow matching
\begin{itemize}
    \item \textbf{Revealing misalignment:} We show that the misalignment between the distribution assumption in regression models and data distribution leads to less accurate estimates and uncalibrated confidence.
    %\item \textbf{Introducing CNFs into Human Pose Estimation:} To the best of our knowledge, this is the first work to apply continuous normalizing flow in Human Pose Estimation. It enables the model to learn the data distribution through continuous-time transformations, thus improving the alignment.
    \item \textbf{Proposing a Novel Method (CFRE):}
    We introduce CNFs into Human Pose Estimation, and to improve the efficiency of model training, we propose a training regime that decouples the CNFs and the regression network.
    CNFs dynamically adjust regression parameters during training without adding extra computation during inference.
    %Moreover, we theoretically show that training with maximum likelihood is equivalent to the flow matching objective with optimal transport vector fields.
    Empirically, our training regime significantly outperforms explicit NLL-based training.
    % \item \textbf{Validation on both 2D and 3D Human Pose Estimation:} Extensive evaluations across MSCOCO \cite{cocodataset} and Human3.6M \cite{h36m_pami} demonstrate that our method outperforms traditional regression models and many heatmap-based methods while producing calibrated confidence. 
    
    %including mean average precision (mAP), area under sparsification error (AUSE), and area under reliability curve (AURG).
\end{itemize}

\vspace{-6mm}

\section{Related Work}

\vspace{-2mm}

\subsection{Heatmap-based Methods}

Heatmap-based approaches were proposed by Tompson \textit{et al.} \cite{NIPS2014_e744f91c}, which utilize likelihood heatmaps to estimate joint locations.
Recently, heatmap-based methods \cite{10.1007/978-3-030-01231-1_29,sun_deep_2019,8237584,10.1007/978-3-030-01264-9_17,mcnally_evopose2d_2021} have dominated due to their ability to model spatial relationships effectively.
Sun \textit{et al.} \cite{sun_deep_2019} achieved competitive performance by maintaining the high resolution of image features during forward propagation in CNNs.
However, these methods suffer from high computational and storage demands, making them unsuitable for scenarios requiring efficiency, such as edge-device deployments.

\vspace{-2mm}

\subsection{Regression-based Methods}

Unlike heatmap-based methods, regression-based approaches \cite{zhou_objects_2019,9010416,10.1007/978-3-030-58607-2_31} prioritize computational efficiency but often lag behind in performance.
To estimate the confidence of model predictions, regression-based methods commonly employ heteroscedastic regression \cite{kendall_what_2017}, which models per-sample variance. 
However, this approach typically assumes a specific distribution, and deviations from the true data distribution can result in suboptimal performance.
Recently, Li et al.~\cite{li_human_2021} proposed leveraging normalizing flows (NFs) to model data distributions without relying on specific assumptions. However, the discrete-time transformations in normalizing flows may limit their ability to accurately capture complex data distributions.
We adopt continuous normalizing flows, which learn such distributions through continuous-time transformations and have demonstrated superior performance in high-dimensional data generation tasks.

\section{Method}

In this section, we introduce a novel regression-based framework that integrates CNFs into human pose estimation. This approach addresses the limitations of fixed distributional assumptions by allowing the model to dynamically learn complex data distributions. Additionally, a specialized loss function is designed to enhance both accuracy and uncertainty quantification. 

\vspace{-2mm}

\subsection{Regression Paradigm}
\label{sec:regression}

The proposed model employs a top-down 2D human pose estimation approach.
%which involves detecting human instances in an image, cropping them into fixed-size sub-images, and predicting the joint locations
For a given image $\mathcal{I}$, the model predicts the coordinates of $K$ joints $\hat{\mathbf{p}} \in \mathbb{R}^{K \times 2}$ and their corresponding joint-wise confidence scores $\hat{s} \in [0,1]^K$. These joint-wise confidence scores are averaged into an instance-wise confidence score $\hat{c} \in [0,1]$, which represents the overall reliability of the model's predictions.

The regression paradigm estimates the deterministic coordinates of joints and corresponding scale by modeling the joint distribution $P_{\beta}(x|\mathcal{I})$ as a probabilistic distribution parameterized by the mean $\hat{\mu}$ and standard deviation $\hat{\sigma}$ using heteroscedastic Deep Regression \cite{kendall_what_2017}. Here, $\beta$ denotes the learnable parameters of the regression model. The distribution is typically assumed to be Gaussian or Laplacian, with the regression model estimating $\hat{\mu}$ and $\hat{\sigma}$ for each joint given an input image $\mathcal{I}$. We follow the previous work \cite{gu_calibration_2023} to normalize $\hat{\sigma}$ to the range $[0,1]$.

To optimize the regression model, the negative log-likelihood (NLL) of the ground truth joint locations $\mu_g$ is minimized as follows:

\begin{equation}
    \mathcal{L}_{\text{reg}} = -\log P_{\beta}(x|\mathcal{I}) \Big|_{x = \mu_g}.
\label{eq:lreg}
\end{equation}

From the distribution parameters, the predicted joint locations $\hat{\mathbf{p}}$ and confidence scores $\hat{s}_k$ for each joint are obtained as:

\begin{equation}
    \hat{\mathbf{p}} = \hat{\mu}, \quad \hat{s}_k = 1 - \hat{\sigma}.
\end{equation}

However, the fixed assumptions about the joint distributions in traditional regression methods limit their flexibility in adapting to the complex nature of human pose data. To address this issue, we propose integrating CNFs into the regression paradigm.

\vspace{-2mm}

\subsection{Reparameterization}
\begin{figure}[h]
    \centering
    \includegraphics[width=0.8\textwidth]{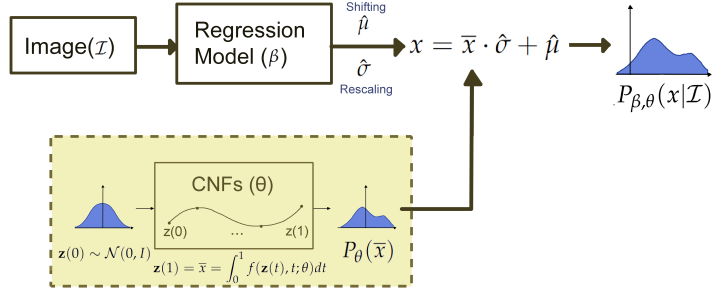}
    \vspace{-5mm}
    \caption{The proposed CFRE architecture. The yellow box is our contribution.}
    \label{fig:cnf_arch}
\end{figure}

To enable seamless integration of CNFs with regression models and to simplify training, we employ a reparameterization technique based on previous works \cite{li_human_2021}. The architecture of the proposed method is illustrated in Figure \ref{fig:cnf_arch}.

Here we assume that the underlying joint distributions belong to the same family of density functions but may differ in their means and standard deviations depending on the input $\mathcal{I}$. The regression model, parameterized by $\beta$, predicts the mean $\hat{\mu}$ and standard deviation $\hat{\sigma}$ of the joint locations. Using these predictions, the CNFs, parameterized by $\theta$, transform a simple initial distribution into a complex normalized distribution.
\( z(t) \) here denotes the state of the transformation at time \( t \), and \( f(z(t), t; \theta) \) defines a velocity field parameterized by \( \theta \). 
The process starts from a simple base distribution \( z(0) \sim \mathcal{N}(0, I) \) and evolves into a target distribution \( z(1) \sim P_\theta(\bar{x})\) by CNFs, that captures the data distribution. 
%This continuous trajectory is termed the \textbf{Flow Map}, enabling smooth transformations between distributions.

The reparameterized process begins by standardizing the ground truth joint location $x$ as follows:
\begin{equation}
    \overline{x} = \frac{x - \hat{\mu}}{\hat{\sigma}},
\end{equation}
where $\overline{x} \sim P_{\theta}(\overline{x})$. Here, $P_{\theta}(\overline{x})$ represents the distribution learned by the CNFs, and $\overline{x}$ is the standardized residual. The original joint location distribution is reconstructed using:
\begin{equation}
    x = \overline{x} \cdot \hat{\sigma} + \hat{\mu} \sim P_{\beta,\theta}(x|\mathcal{I}),
\end{equation}
where $P_{\beta,\theta}(x|\mathcal{I})$ is the final predicted distribution of the joint location.

\vspace{-2mm}

\subsection{Negative Log-Likelihood Estimation with CNFs}
\label{sec:nll_cnf}

An intuitive way to train the parameters of both the regression model and CNFs is to minimize the negative log-likelihood (NLL) of the predicted joint distribution \( P_{\beta,\theta}(x|\mathcal{I}) \). The reparameterization introduced earlier allows us to express the NLL as a function of the initial distribution and the CNF transformation dynamics.

\vspace{-5mm}

\subsubsection{Density Transformation via CNFs}
Using the continuous transformation defined by CNFs, the log-probability of the normalized variable \( \overline{x} = \frac{x - \hat{\mu}}{\hat{\sigma}} \), where \( \overline{x} \sim P_{\theta}(\overline{x}) \), evolves as follows \cite{NEURIPS2018_69386f6b}:
\begin{equation}
    \frac{\partial \log P_{\theta}(z(t))}{\partial t} = -\text{tr} \left( \frac{\partial f(z(t), t; \theta)}{\partial z(t)} \right),
\end{equation}
where \( f(z(t), t; \theta) \) represents the vector field parameterized by \( \theta \), and \( \text{tr}(\cdot) \) denotes the trace operator. By integrating this differential equation over the interval \( [0, 1] \), we obtain:
\begin{equation}
    \log P_{\theta}(\overline{x}) = \log P_{\theta}(z(0)) - \int_0^1 \text{tr} \left( \frac{\partial f(z(t), t; \theta)}{\partial z(t)} \right) dt.
\end{equation}

The full joint distribution \( P_{\beta,\theta}(x|\mathcal{I}) \) is then reconstructed by applying the change-of-variable formula:
\begin{equation}
    P_{\beta,\theta}(x|\mathcal{I}) = \frac{1}{\hat{\sigma}} \cdot P_{\theta} \left( \frac{x - \hat{\mu}}{\hat{\sigma}} \right).
\end{equation}

\subsubsection{Optimizing the NLL}
By taking the negative log of \( P_{\beta,\theta}(x|\mathcal{I}) \), we obtain the loss function that we use for training:
\begin{equation}
    \mathcal{L}_{\text{NLL}} = \log \hat{\sigma} - \log P_{\theta}(z(0)) + \int_0^1 \text{tr} \left( \frac{\partial f(z(t), t; \theta)}{\partial z(t)} \right) dt.
\end{equation}

The key computational challenge lies in evaluating the trace term, which involves the Jacobian of the vector field and is computationally expensive.

\vspace{-5mm}

\subsubsection{Trace Estimation with the Hutchinson Trace Estimator (HTE).}
To mitigate the computational burden, we use the Hutchinson estimator \cite{grathwohl_ffjord_2018}, which approximates the trace of a matrix through stochastic sampling:
\begin{equation}
    \text{tr} \left( \frac{\partial f(z(t), t; \theta)}{\partial z(t)} \right) \approx \mathbb{E}_{\varepsilon \sim \mathcal{N}(0, I)} \left[ \varepsilon^\top \frac{\partial f(z(t), t; \theta)}{\partial z(t)} \varepsilon \right],
\end{equation}
where \( \varepsilon \) is a random vector sampled from a standard normal distribution.

\vspace{-2mm}

\subsection{Training of the Model}
\label{sec:training_model}

The proposed training approach optimizes both the regression network and Continuous Normalizing Flows (CNFs) in an end-to-end manner. However, while the HTE efficiently approximates the trace of the Jacobian matrix, it introduces variance. This can lead to non-smooth loss curves and convergence difficulties. In addition, the numerical integration may also introduce errors.

% Additionally, the regression network relies heavily on distributions estimated by CNFs. When CNFs are not yet well-trained, the regression network learns from an incorrect distribution, leading to suboptimal performance.

\vspace{-5mm}

\subsubsection{Decoupled Training Strategy}
To address these issues, we propose a decoupled training strategy of the regression network and CNFs. The total loss function is decomposed as:
\begin{equation}
    \mathcal{L}_{\text{total}} = \mathcal{L}_{\text{reg}} + \lambda(\hat{\sigma}) \cdot \mathcal{L}_{\text{flow}},
    \label{equ:cfre_loss}
\end{equation}
where 
\begin{equation}
\label{equ:hyper_c}
    \lambda(\hat{\sigma}) = c \cdot (1 - \hat{\sigma}),
\end{equation}
is the self-adaptive scaling factor and \( c \) is a hyperparameter.
Here \( \mathcal{L}_{\text{reg}} \) trains the regression network to fit a reference distribution, while \( \mathcal{L}_{\text{flow}} \) refines the reference by modeling residuals via CNFs and enables the regression network to adapt to the new data distribution during training, leading to more precise scale estimation.\\
\parsection{Self-adaptive scaling factor}  \( \lambda(\hat{\sigma}) \) adjusts the influence of CNFs component based on the uncertainty, \( \hat{\sigma} \), of the predictions of the regression network.
We notice that during the early training epochs, CNFs are not yet well-trained, hence downweighting by the self-adaptive scaling factor will enable the regression network to learn robust initial estimates, while CNFs refine these estimates without imposing strong dependencies during early training stages. Furthermore, the dynamic weighting ensures effective training across samples with varying uncertainties, leading to a more robust model.

\subsubsection{Continuous Normalizing Flows Training}
To avoid explicitly calculating the divergence of the vector field on $z(t)$ (i.e., the trace of the Jacobian matrix) and integration, we optimize an upper bound to simplify the training:
\begin{equation}
    \mathcal{L}_{\text{UB}} = -\log P(z(0)) + n \sup_{z(t)} \left\| \frac{\partial f(z(t), t; \theta)}{\partial z(t)} \right\|,
\end{equation}
where \( n \) is the dimension of \( z(t) \), and \( \sup_{z(t)} \| \cdot \| \) denotes the Lipschitz constant of \( f \).
Inspired by Flow Matching \cite{lipman_flow_2023}, we specify an optimal transport path \( z(t) \) between the initial and target distributions:
\begin{equation}
    z(t) = (1 - (1 - \sigma_{\min})t)z(0) + tz(1),
\end{equation}
where \( \sigma_{\min} \) is the standard deviation of the target distribution. The corresponding vector field is:
\begin{equation}
    u(z(t), t) = \frac{z(1) - (1 - \sigma_{\min})z(t)}{1 - (1 - \sigma_{\min})t}.
\end{equation}

By training \( f(z(t), t; \theta) \) to match \( u(z(t), t) \), the term \( \sup_{z(t)} \left\| \frac{\partial f(z(t), t; \theta)}{\partial z(t)} \right\| \) approaches 0, meaning that the upper bound \( \mathcal{L}_{\text{UB}} \) approaches its minimum value.
Hence the $\mathcal{L}_\textup{flow}$ is given by (see also Lipman \textit{et al.}~\cite{lipman_flow_2023}):
\begin{equation}
    \mathcal{L}_\textup{flow} = E_{t,q(\overline{x}),p(z(0))}||f(z(t),t;\theta)-(\overline{x}-(1-\sigma_{\min})z(0))||^{2}
\end{equation}

\vspace{-8mm}

\subsubsection{Heteroscedastic Regression Network Training}

The regression network models the target distribution using a Laplace distribution as the reference distribution, characterized by a mean, $\hat{\mu}$  and a scale parameter, $\hat{\sigma}$. This results in the following expression for \eqref{eq:lreg}:

\begin{equation}
    \mathcal{L}_\textup{reg}=\log(\sqrt{2}\hat{\sigma})+\frac{2|\mu_g-\hat{\mu}|}{\sqrt{2}\hat{\sigma}}
\end{equation}

\vspace{-5mm}

\subsection{Evaluation Metrics}
% cite human pose calibartion, indicating better confidence ranking leads to better mAP. Evaluate this with AUSE, AURG.

For 2D and 3D HPE, the primary evaluation metric is mean Average Precision (mAP).
It depends not only on the localization error but also on the ranking of predictive uncertainty \cite{gu_calibration_2023}.

When predictive uncertainty is consistent with localization error in ranking, this leads to a higher mAP.
To further validate the improvement on predictive uncertainty, we adopt Area under Sparsification Error (AUSE) from \cite{ilg2018uncertainty} and Area under Reliability Curve (AURG) \cite{poggi_uncertainty_2020}. 
In \cite{lind2024uncertainty} Lind et al.\ demonstrate that AUSE is more robust than ranking-based UQ metrics for regression tasks.

\vspace{-5mm}

\section{Experiments}
\vspace{-3mm}

We evaluate our proposed methods on both 2D human pose in Sec.~\ref{sec:coco} and 3D human pose tasks Sec.~\ref{sec:human3.6m}.

%\vspace{-7mm}

\subsection{Experiments on COCO}
\label{sec:coco}
\vspace{-2.5mm}

To demonstrate the effectiveness of our proposed method, we compare CFRE against heatmap-based methods, and state-of-the-art approaches using the COCO dataset.
Our results consistently highlight the advantages of CFRE in terms of accuracy, uncertainty quantification, and computational efficiency.

\parsection{Dataset} We first evaluate the proposed method on a large-scale in-the-wild 2D human pose benchmark COCO-Pose \cite{cocodataset}.
This dataset provides annotations of 17 keypoints for body parts like the nose, eyes, and shoulders. 
It includes diverse human poses in various environments, making it a challenging yet realistic dataset. 
COCO is split into training, validation, and test-dev sets, with the latter excluding bounding box annotations to ensure fair evaluation and simulate real-world conditions.

\parsection{Implementation Details}
The proposed model, CFRE, is validated on the COCO dataset using a top-down approach described in \cite{li_human_2021}. 
%This involves detecting human instances in images followed by pose estimation using CFRE.
%For human instance detection, we employ the person detectors provided by SimplePose \cite{10.1007/978-3-030-01231-1_29}.
During inference, only the regression network within CFRE is utilized.
The regression network consists of a backbone, an average pooling layer, and a fully connected (FC) layer.
It predicts the estimated mean joint locations \( \hat{\mu} \), and the standard deviation \( \hat{\sigma} \) of joint locations along both axes.
The output dimension is \( K \times 4 \), where \( K = 17 \) corresponds to the number of keypoints, and 4 represents \( \hat{\mu} \) and \( \hat{\sigma} \).

To enhance generalization, we adopt the data augmentation techniques commonly used in prior works \cite{sun_deep_2019}.
The loss balancing weight \( c \) is set to \( 0.1 \), based on  cross validation, see plot in Figure\ \ref{fig:hyper_c_chart}.  
%The weighting factor \( c \) in the loss function is set to \( 0.1 \), balancing the contributions of regression and flow components effectively.
%These include scale and rotation augmentation to handle targets of varying sizes and orientations, random horizontal flipping to reduce directional bias, and half-body transformation, which randomly crops the upper or lower body of human instances with a predefined probability, effectively enriching the training dataset with partial-body views. Additionally, affine transformation is applied to normalize input images to a fixed size, improving the model's adaptability to variations in shape and size.

%Training is performed for 300 epochs using 4 NVIDIA A100 GPUs. 

For experiments on the COCO validation set, the COCO train set is divided into 90\% training and 10\% validation subsets, with the COCO val set used as the test set. For evaluations on the COCO test-dev set, the entire COCO train set serves as the training set, the COCO val set is used for validation, and the COCO test-dev set is employed as the test set.

\textbf{Comparison with the SotA.}  
We compare CFRE with SOTA single-stage and two-stage methods on the COCO val and test-dev sets. We use 'ResNet-152 + FC Layer' as the architecture for our regression model, with input image resolution set to $384 \times 233$. As shown in Tables \ref{tab:pose_comparison_sota_val} and \ref{tab:pose_comparison_sota_test}, CFRE achieves competitive mAP among regression-based and heatmap-based methods. 
We alse evaluate computational efficiency in terms of GFLOPs, where lower values indicate higher efficiency. As shown in Table \ref{tab:pose_comparison_sota_val}, CFRE achieves competitive performance, outperforming SimplePose with the same backbone and approaching HRNet's performance, while requiring significantly fewer computational resources.
%\vspace{-5mm}

\begin{figure}[htbp] 
    \centering 
    \includegraphics[width=1.0\textwidth]{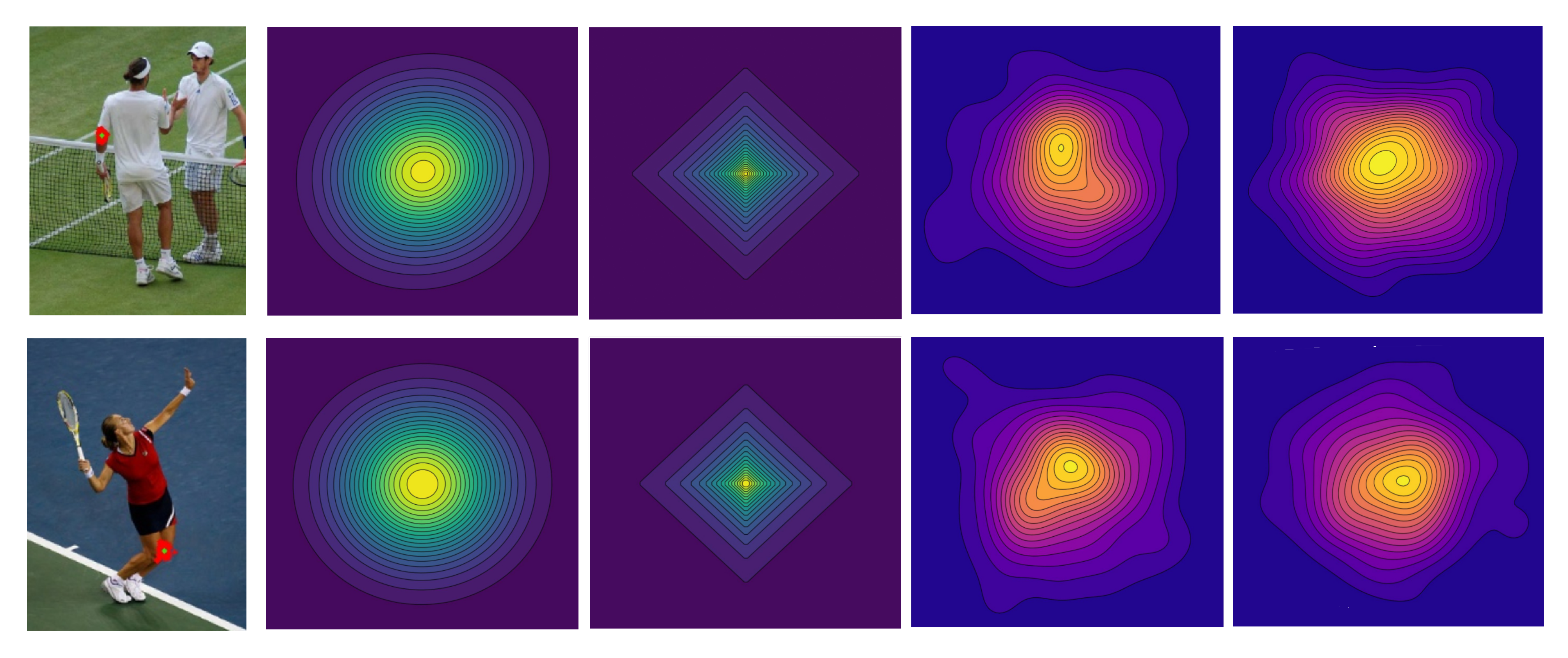} 
    \caption{Contour plot of joint estimations across multiple human instances. \textbf{First column:} Input images (Red: Samples; Green: Mean value); \textbf{Second column:} Estimation with Gaussian assumption; \textbf{Third column:} Estimation with Laplace assumption; \textbf{Fourth column:} Distributions learned by CNFs. \textbf{Fifth column:} Distributions learned by NFs. Rows correspond to different human instances and their joint estimations: (1) \textbf{left elbow}, (2) \textbf{right knee}.} 
    \label{fig:contour_plot} 
    \vspace{-5mm}
\end{figure}

\textbf{Qualitative Result on COCO.}  
To compare the learned distributions qualitatively, CNFs were used to sample the model's joint location estimates, visualized as contour plots. The result is shown in Figure \ref{fig:contour_plot}.
Our proposed model was compared against heteroscedastic regression models assuming Laplace and Gaussian distributions, with CFRE sampled 200 times.
The learned distribution's peak is broader than the Laplace distribution but sharper than the Gaussian, with heavier tails. 
The vertical variance exceeds the horizontal, indicating asymmetric spread.
Moreover, it can be observed from the figure that CNFs capture more complex distributions compared to NFs.

\vspace{-5mm}

\begin{table}[H]
\centering
    \caption{Comparison of different methods on COCO val.}
    \label{tab:pose_comparison_sota_val}
\begin{tabular}{llllllll}
\hline
\multicolumn{1}{l|}{Method}              & \multicolumn{1}{l|}{Backbone}    & $mAP$         & $AP_{50}$     & $AP_{75}$     & $AP_M$        & $AP_L$        & GFLOPs \\ \hline
\textit{Heatmap-based}                   &                                  &               &               &               &               &               &        \\ \hline
\multicolumn{1}{l|}{Mask R-CNN \cite{8237584}}          & \multicolumn{1}{l|}{Resnet-101}  & 66.1          & 87.4          & 72.0          & 61.5          & 74.4          & -      \\
\multicolumn{1}{l|}{PifPaf \cite{Kreiss2019PifPafCF}}              & \multicolumn{1}{l|}{ResNet-152}  & 67.4          & -             & -             & -             & -             & -      \\
\multicolumn{1}{l|}{PersonLab \cite{10.1007/978-3-030-01264-9_17}}           & \multicolumn{1}{l|}{ResNet-152}  & 66.5          & 86.2          & 71.9          & 62.3          & 73.2          & 405.5      \\
\multicolumn{1}{l|}{AE \cite{NIPS2017_8edd7215}}                  & \multicolumn{1}{l|}{HrHRNet-W48} & 72.1          & 88.4          & 78.2          & -             & -             & -      \\
\multicolumn{1}{l|}{SimplePose \cite{10.1007/978-3-030-01231-1_29}}          & \multicolumn{1}{l|}{Resnet-152}  & 74.3          & 89.6          & 81.1          & 70.5          & 81.6          & 35.3      \\
\multicolumn{1}{l|}{HRNet \cite{sun_deep_2019}}               & \multicolumn{1}{l|}{HRNet-W48}   & \textbf{76.3} & \textbf{90.8} & \textbf{82.9} & \textbf{72.3} & \textbf{83.4} & \textbf{32.9}      \\ \hline
\textit{Regression-based}                &                                  &               &               &               &               &               &        \\ \hline
\multicolumn{1}{l|}{DERK \cite{9578170}}                & \multicolumn{1}{l|}{HRNet-W32}   & 67.2          & 86.3          & 73.8          & 61.7          & 77.1          & 45.4      \\
\multicolumn{1}{l|}{PRTR \cite{Li_2021_CVPR}}                & \multicolumn{1}{l|}{HRNet-W32}   & 73.3          & 89.2          & 79.9          & 69.0          & \textbf{80.9} & 37.8      \\
\multicolumn{1}{l|}{RLE \cite{li_human_2021}}                 & \multicolumn{1}{l|}{Resnet-152}  & 75.4          & 91.3          & 82.0          & \textbf{75.1} & 78.4          & 24.9      \\
\multicolumn{1}{l|}{\textbf{CFRE(Our)}} & \multicolumn{1}{l|}{Resnet-152}  & \textbf{75.6} & \textbf{92.2} & \textbf{82.1} & 75.0          & 78.5          & \textbf{24.9}      \\ \hline
\end{tabular}
\vspace{-5mm}
\end{table}

\begin{table}[H]
\vspace{-5mm}
\centering
    \caption{Comparison of different methods on COCO test-dev.}
    \label{tab:pose_comparison_sota_test}

\begin{tabular}{lllllll}
\hline
\multicolumn{1}{l|}{Method}              & \multicolumn{1}{l|}{Backbone}         & $mAP$ & $AP_{50}$ & $AP_{75}$ & $AP_M$ & $AP_L$ \\ \hline
\textit{Heatmap-based}                   &                                       &       &           &           &        &        \\ \hline
\multicolumn{1}{l|}{CMU-Pose \cite{8765346}}            & \multicolumn{1}{l|}{3CM-3PAF}         & 61.8  & 84.9      & 67.5      & 57.1   & 68.2   \\
\multicolumn{1}{l|}{Mask R-CNN \cite{8237584}}          & \multicolumn{1}{l|}{Resnet-50}        & 63.1  & 87.3      & 68.7      & 57.8   & 71.4   \\
\multicolumn{1}{l|}{RMPE \cite{8237518}}                & \multicolumn{1}{l|}{PyraNet}          & 72.3  & 89.2      & 79.1      & 68.0   & 78.6   \\
\multicolumn{1}{l|}{AE \cite{NIPS2017_8edd7215}}                  & \multicolumn{1}{l|}{Hourglass-4}      & 65.5  & 86.8      & 72.3      & 60.6   & 72.6   \\
\multicolumn{1}{l|}{PersonLab \cite{10.1007/978-3-030-01264-9_17}}           & \multicolumn{1}{l|}{Resnet-152}       & 68.7  & 89.0      & 75.4      & 64.1   & 75.5   \\
\multicolumn{1}{l|}{SimplePose \cite{10.1007/978-3-030-01231-1_29}}          & \multicolumn{1}{l|}{Resnet-152}       & 73.7  & 91.9      & 81.1      & 70.3   & 80.0   \\
\multicolumn{1}{l|}{Integral \cite{10.1007/978-3-030-01231-1_33}}            & \multicolumn{1}{l|}{Resnet-101}       & 67.8  & 88.2      & 74.8      & 63.9   & 74.0   \\
\multicolumn{1}{l|}{HRNet \cite{sun_deep_2019}}               & \multicolumn{1}{l|}{HRNet-W48}        & 75.5  & {\bf 92.5}      & {\bf 83.3}      & 71.9   & {\bf 81.5}   \\
\multicolumn{1}{l|}{EvoPose \cite{mcnally_evopose2d_2021}}             & \multicolumn{1}{l|}{EvoPose2D-L}      & {\bf 75.7}  & 91.9      & 83.1      & {\bf 72.2}   & {\bf 81.5}   \\ \hline
\textit{Regression-based}                &                                       &       &           &           &        &        \\ \hline
\multicolumn{1}{l|}{CenterNet \cite{zhou_objects_2019}}           & \multicolumn{1}{l|}{Hourglass-2}      & 63.0  & 86.8      & 69.6      & 58.9   & 70.4   \\
\multicolumn{1}{l|}{SPM \cite{9010416}}                 & \multicolumn{1}{l|}{Hourglass-8}      & 66.9  & 88.5      & 72.9      & 62.6   & 73.1   \\
\multicolumn{1}{l|}{PointSet Anchor \cite{10.1007/978-3-030-58607-2_31}}     & \multicolumn{1}{l|}{HRNet-W48}        & 68.7  & {\bf 89.9}      & 76.3      & 64.8   & 75.3   \\
\multicolumn{1}{l|}{RLE \cite{li_human_2021}}                 & \multicolumn{1}{l|}{Resnet-152}       & {\bf 74.2}  & 89.5      & {\bf 80.7}      & {\bf 71.0}   & 79.7   \\
\multicolumn{1}{l|}{\textbf{CFRE(Our)}} & \multicolumn{1}{l|}{Resnet-152}       & 73.8  & 89.1      & 80.4      & 70.4   & {\bf 80.0}   \\ \hline
\end{tabular}
\end{table}

\subsection{Ablation Study}
\label{sec:abs}
Firstly, we train heteroscedastic regression models only with the assumption of different distributions.
We evaluate the proposed training paradigm in Sec.\ \ref{sec:training_model} and compare it with  the explicit NLL training in Sec.\ \ref{sec:nll_cnf}.
We also investigate the impact of the hyperparameter \( c \), which balances the CFRE  loss \eqref{equ:cfre_loss}.

\begin{figure}[ht]
    \centering
    \begin{minipage}[b]{0.45\textwidth}
        \centering
        \includegraphics[width=\textwidth]{figures/hyper_c_chart.pdf}
        \caption{Impact of Hyperparameter $c$ on AP metrics.}
        \label{fig:hyper_c_chart}
    \end{minipage}
    \hfill
    \begin{minipage}[b]{0.45\textwidth}
        \centering
        \includegraphics[width=\textwidth]{figures/sparsification_error.pdf}
        \caption{Sparsification Error Plots Comparing Different Regression Methods.}
        \label{fig:spar_fig}
    \end{minipage}
\end{figure}

\parsection{Comparison with Heteroscedastic Regression.}  
We compare CFRE with heteroscedastic deep regression models assuming Gaussian and Laplace distributions. All models use the proposed architecture in Sec.\ \ref{sec:coco}, with the input size set to $384 \times 288$, and are trained on the COCO dataset.
\textbf{Results:} As shown in Table \ref{tab:reg_combined}, Laplace-based training loss outperforms Gaussian based by $3\%$ in mAP and lower AUSE.
It suggests a better alignment between the distribution assumption and data distirbution will improve performance. 
Furthermore, CFRE achieves a 2.4 mAP improvement over the Laplace-based model, outperforming both heteroscedastic regression models in uncertainty quantification metrics.
This further validates the flexibility of our proposed architecture. 

\vspace{-5mm}

\begin{table}[h]
    \centering
    \caption{Comparison with Heteroscedastic Regression on mAP and Uncertainty Metrics.}
    \label{tab:reg_combined}
    \begin{tabular}{lllllll}
    \hline
    Method                                        & Backbone      & mAP $\uparrow$        & PCC $\uparrow$ & AUSE $\downarrow$ & AURG $\uparrow$ \\ \hline
    heteroscedastic Regression (Gaussian)        & ResNet-152 & 70.7          & 0.476          & 0.225          & 0.273          \\
    heteroscedastic Regression (Laplace)         & ResNet-152 & 73.2          & 0.522          & 0.221          & 0.280          \\
    \textbf{CFRE(Our)}                         & ResNet-152 & \textbf{75.6} & \textbf{0.556} & \textbf{0.195} & \textbf{0.441} \\ \hline
    \end{tabular}
    \vspace{-5mm}
\end{table}

\vspace{+5mm}

\parsection{Comparison with Explicit NLL Optimization.}  
The baseline method directly optimizes the explicit NLL formula by estimating the trace of the Jacobian matrix using the Hutchinson estimator. In contrast, CFRE leverages Flow Matching with Optimal Transport Conditional Vector Fields, bypassing the need for Jacobian trace computation and numerical integration during training.

In Table \ref{tab:pose_comparison}, we compare the two approaches on the COCO val. CFRE consistently outperforms the baseline in mAP, demonstrating the effectiveness of bypassing computational bottlenecks while maintaining high accuracy.

\begin{table}[h]
    \centering
    \caption{Performance Comparison of training with Explicit NLL and CFRE.}
    \label{tab:pose_comparison}
    \begin{tabular}{llllll}
    \hline
    Method  & Backbone   & Input Size   & $\text{mAP}$ & $\text{AP}_{50}$ & $\text{AP}_{75}$ \\ \hline
    Regression with CNF (Explicit NLL)      & ResNet-50  & $256 \times 192$ & 68.0  & 88.7      & 75.1      \\
    \textbf{CFRE(Our)}                    & ResNet-50  & $256 \times 192$ & \textbf{70.7}  & \textbf{90.5}      & \textbf{77.1}      \\
    Regression with CNF (Explicit NLL)      & ResNet-152 & $384 \times 288$ & 73.8  & 89.1      & 80.0      \\
    
    \textbf{CFRE(Our)}                    & ResNet-152 & $384 \times 288$ & \textbf{75.6}  & \textbf{92.2}      & \textbf{82.2}      \\ \hline
    \end{tabular}
\end{table}

\parsection{Effect of Hyperparameter \( c \).}  
The hyperparameter \( c \) in Eq.\ (\ref{equ:hyper_c}) determines the balance between the regression loss \( \mathcal{L}_{\text{reg}} \) and the CNF loss \( \mathcal{L}_{\text{flow}} \). Figure \ref{fig:hyper_c_chart} illustrates the impact of \( c \) on the COCO validation set. The model achieves peak performance when \( c = 0.1 \) on COCO val. Smaller values of \( c \) (e.g., \( c = 0.0 \)) reduce the contribution of \( \mathcal{L}_{\text{flow}} \), degrading the model to heteroscedastic regression with Laplace assumption.
% \begin{figure}[h]
%     \centering
%     \includegraphics[width=0.7\textwidth]{figures/hyper_c_chart.png}
%     \caption{Impact of Hyperparameter $c$ on Performance}
%     \label{fig:hyper_c_chart}
% \end{figure}
This study confirms that balancing the loss components through \( c \) is critical to performance, with \( c = 0.1 \) providing the best trade-off between regression accuracy and distribution refinement.

\vspace{-2mm}

\subsection{Experiments on Human3.6M}
\label{sec:human3.6m}

We further evaluate our method on the Human3.6M dataset \cite{IonescuSminchisescu11,h36m_pami}, a large-scale benchmark designed for 3D human pose estimation. 
For evaluation, we report results using metrics like Mean Per Joint Position Error (MPJPE) and Procrustes-Aligned MPJPE (PA-MPJPE) under standardized protocols. In additioin, we use AUSE to report the uncertainty quantification performance of the models for 3D pose estimation tasks.

\parsection{Implementation Details} For the experiments conducted on the Human3.6M dataset, we adopted a single-stage approach, following the training and testing procedure of previous work \cite{li_human_2021}.
The architecture used was the same as ResNet-50 with a FC layer. 
Input images were normalized to a size of $256 \times 256$, and data augmentation techniques were consistent with those in previous research \cite{li_human_2021}. 
The initial learning rate was set to $1 \times 10^{-3}$ and was decayed to $1/10$ at Epochs 120 and 170. 
The model was trained for 200 epochs using the Adam optimizer, with a mini-batch size of 32 per GPU and a total of 2 GPUs.

\begin{table}[h]
    \centering
    \caption{Comparison on Human3.6M}
    \label{tab:single_stage}
    \begin{tabular}{lccccc}
    \hline
    Method & \#Params & GFLOPs & MPJPE $\downarrow$ & PA-MPJPE $\downarrow$ & AUSE $\downarrow$ \\ \hline
    Direct Regression & 23.8M & 5.4 & 50.1 & 39.3 & 0.279\\
    
    Heteroscedastic Reg (Laplace) & 23.8M & 5.4 & 50.8 & 39.7 & 0.112\\
    %Integral Pose \cite{10.1007/978-3-030-01231-1_33} & 34.3M & 14.1 & 49.2 & 39.1 & 0.0\\
    RLE \cite{li_human_2021} & 23.8M & 5.4 & 48.6 & 38.5 & 0.106\\
    \textbf{Regression with CFRE} & \textbf{23.8M} & \textbf{5.4} & \textbf{48.2} & \textbf{38.2} & \textbf{0.101}\\ \hline
    \end{tabular}
    \vspace{-5mm}
\end{table}

\parsection{Results} Table \ref{tab:single_stage} highlights the improvements achieved by our method on the Human3.6M dataset. 
Our approach achieves the lowest MPJPE (48.2 mm) and PA-MPJPE (38.2 mm), while maintaining the same GFLOPs (5.4) as RLE and heteroscedastic regression.
In addition, our method surpasses the baseline heteroscedastic regression by 1.1\% and RLE by 0.5\% on AUSE.
%The sparcification plots of heteroscedastic regression, regression with RLE and CFRE are as shown in Figure \ref{fig:spar_fig}.

The sparsification error plots of heteroscedastic regression, regression with RLE and CFRE are as shown in Figure \ref{fig:spar_fig}.
Sparsification error is the difference between the model sparsification curve and the oracle curve, indicating the gap between the model and the optimal uncertainty, see \cite{lind2024uncertainty}.

\vspace{-5mm}

% \begin{figure}[h]
%     \centering
%     %\subfloat[Direct Regression]{\includegraphics[width=0.33\textwidth]{figures/sparsification_direct_reg.png}}
%     \subfloat[Heteroscedastic Regression (Laplace)]{\includegraphics[width=0.33\textwidth]{figures/sparsification1_hetero.png}}
%     \subfloat[Regression with RLE]{\includegraphics[width=0.33\textwidth]{figures/sparsification1_rle.png}}
%     \subfloat[Regression with CFRE]{\includegraphics[width=0.33\textwidth]{figures/sparsification1_cfre.png}}
%     \caption{Sparsification Plots Comparing Different Regression Methods}
%     \label{fig:spar_fig}
% \end{figure}

\section{Conclusion}

\vspace{-2mm}

This study enhances regression-based Human Pose Estimation by integrating CNFs. This is shown to improve accuracy and uncertainty quantification without increasing computational costs.
CNFs allow the model to dynamically learn joint location distributions, overcoming the limitations of fixed assumptions.
Analysis of the learned distributions reveals sharper peaks, heavier tails, and axis-specific variances, aligning with real-world data characteristics.

The proposed model demonstrates superior performance on the COCO and Human3.6M dataset, achieving notable gains in evaluation metrics related to uncertainty such as mAP, Pearson correlation coefficient (PCC) and AUSE over traditional regression models while maintaining computational efficiency.
It surpasses state-of-the-art regression methods and outperforms heatmap-based approaches with the same backbone, delivering improved accuracy at a lower computational cost.
To optimize training, we derive an upper bound for the NLL of CNFs, proving its equivalence to Flow Matching training with Optimal Transport vector fields.

In summary, this work establishes CNFs as a powerful enhancement to regress-ion-based HPE, achieving state-of-the-art results while maintaining efficiency.

\begin{credits}
\subsubsection{\ackname} This work was supported by the strategic research environment ELLIIT funded by the Swedish government, and by Vinnova, project \textnumero $2023-02694$, and Formas project \textnumero 2023-00082. Computational resources were provided by the National Academic Infrastructure for Supercomputing in Sweden (NAISS), partially funded by the Swedish Research Council through grant agreement no. 2022-06725.
%\subsubsection{\ackname} Placeholder.

%\subsubsection{\discintname}
%The authors have no competing interests to declare that are relevant to the content of this article.
\end{credits}

\bibliographystyle{splncs04}
\bibliography{mybibliography}

\title{Supplementary Material for the Paper: Continuous Normalizing Flows for Uncertainty-Aware Human Pose Estimation}
%
%\titlerunning{Abbreviated paper title}
% If the paper title is too long for the running head, you can set
% an abbreviated paper title here
%
\author{ }
%\author{Shipeng Liu\inst{1}\orcidID{0000-1111-2222-3333} \and Ziliang Xiong\inst{1}\orcidID{0009-0008-8277-7476} \and  Bastian Wandt\inst{1}\orcidID{2222--3333-4444-5555} \and  Per-Erik Forssén\inst{1}\orcidID{0000-0002-5698-5983}}
%
\authorrunning{Shipeng Liu et al.}
% First names are abbreviated in the running head.
% If there are more than two authors, 'et al.' is used.
%
\institute{ }
\maketitle              % typeset the header of the contribution
\vspace{-8mm}
\section{Theorem Proof}

\noindent \textbf{Theorem 1}: For an \( n \times n \) matrix \( A \), the inequality
    $\textup{tr}(A) \leq \sqrt{n} \|A\|_F$
holds, where \( \|A\|_F \) is the Frobenius norm of the matrix \( A \).

\vspace{1em}

\textbf{Proof:} 
%From the Cauchy-Schwarz inequality, we have
%\begin{equation}
%    \left( \sum_{i=1}^n A_{ii} \right)^2 \leq n \sum_{i=1}^n A_{ii}^2,
%\end{equation} thus,
\begin{align}
    \textup{tr}(A) &= \sum_{i=1}^n A_{ii} \leq \sqrt{n} \left( \sum_{i=1}^n A_{ii}^2 \right)^{\frac{1}{2}} \text{(Cauchy-Schwarz inequality)}\\
    &\leq \sqrt{n} \left( \sum_{i=1}^n \sum_{j=1}^n |A_{ij}|^2 \right)^{\frac{1}{2}} \\
    &= \sqrt{n} \|A\|_F.
\end{align}

\vspace{1em}

\noindent \textbf{Theorem 2}: For an $n \times n$ matrix $A$, the inequality $ \|A\|_F \leq \sqrt{n} \|A\|$
holds, where $\|A\|_F$ is the Frobenius norm of the matrix $A$ and $\|A\|$ is the spectral norm of this matrix.

\vspace{1em}

\textbf{Proof:} 
\begin{equation}
    \| A \| =\max_{\| x \|_2=1}\| Ax\|_2 =\sigma_\textup{max}(A)=\max_{1 \leq i \leq n} \sigma_i,
\end{equation} where $\sigma_\textup{max}(A)$ represents the largest singular value of $A$, $\sigma_i(A)$ are the singular values of $A$.
Frobenius norm of $A$:
$\| A\|_F=\sqrt{\textup{tr}(A^*A)}=\sqrt{\sum^n_{i=1}\sigma^2_i(A)}.$

Since $\sigma_i \leq \sigma_\textup{max}(A)$ for all $i$, we have $
    \sum^n_{i=1}\sigma_i^2 \leq \sum^n_{i=1} \sigma_\textup{max}(A)^2 =n \cdot \sigma_\textup{max}(A)^2 = n \|A\|^.$

Hence, $\|A\|_F \leq \sqrt{n} \|A\|.$

\vspace{1em}

\noindent \textbf{Theorem 3}: Consider a function $f(z): \mathbb{R}^n \rightarrow \mathbb{R}^n$ with partial derivatives of $z$ everywhere so that the Jacobian matrix is well-defined. Assume $f(z)$ is \textbf{Lipschitz continuous} on $z$. Let $L \geq 0$ be it's \textbf{Lipschitz constant}. Then $
\left\| \frac{\partial f}{\partial z} \right\| \leq L, \forall z,$ 
where $\| \cdot \|$ denotes the spectral matrix norm.

\vspace{1em}

\textbf{Proof:} 

For a fixed \( \delta > 0 \), \( z \in \mathbb{R}^n \), and assuming that the derivative \( \frac{\partial f}{\partial z} \) exists, let \( L \) be the Lipschitz constant of \( f \). Hence, \( f \) is \( (L + \delta)\)-Lipschitz continuous.
Let \( \forall v \in \mathbb{R}^n \) with \( \| v \| = 1 \) be a unit vector. Consider the quantity:

\begin{equation}
    \frac{\left| f(x) - f(x + tv) \right|}{t},
\end{equation}

where \( t \in \mathbb{R} \). When \( t \rightarrow 0 \), \( J_f(x)(v) \) is the directional derivative of \( f \) in the direction \( v \). Thus, there exists \( \epsilon_v > 0 \) such that:

\begin{equation}
    \frac{\left| f(x) - f(x + tv) \right|}{t} < L + \delta, \quad \text{for} \quad |t| < \epsilon_v.
\end{equation}

By the compactness of the unit sphere, there exists a common \( \epsilon > 0 \) such that

\begin{equation}
    \frac{| f(x) - f(z) |}{|x - z|} < L + \delta, \quad \text{if} \quad |x - z| < \epsilon.
\end{equation}

Consider the line segment connecting the points \( x \) and \( y \), covered and divided into a finite number of \( N \) small intervals by a series of balls \( \{B_1, \ldots, B_N\} \), where \( f \) is \( (L + \delta) \)-Lipschitz continuous on each ball, and the centers of these balls lie on the line segment. Adjacent balls intersect each other, so for the center point \( x_i \) of the \( i \)-th ball, we have

\begin{equation}
    |f(x_{i+1}) - f(x_i)| \leq (L + \delta) |x_{i+1} - x_i|.
\end{equation}

Applying the triangle inequality,

\begin{align}
    |f(x) - f(y)| &= |f(x_N) - f(x_1)| \\
    &\leq \sum_{i=1}^N |f(x_{i+1}) - f(x_i)| \\
    &\leq (L + \delta) \sum_{i=1}^N |x_{i+1} - x_i| \\
    &= (L + \delta) |x - y|.
\end{align}

By using more balls to cover the line segment, we can let \( \delta \rightarrow 0 \), so

\begin{equation}
    J_f(x)(v) = \lim_{t \to 0} \frac{\left| f(x) - f(x + tv) \right|}{t} \leq L,
\end{equation} thus,

\begin{equation}
    \left\| \frac{\partial f}{\partial z} \right\| = \left\| J_f(x) \right\| = \sup_{\substack{v \in \mathbb{R}^n \\ \|v\| = 1}} J_f(x)(v) \leq L.
\end{equation}

This result is trivial when $n=1$. we can use the definition of the derivative and the mean value theorem to obtain the conclusion of the theorem 3. In the proof above, we have extended it to higher dimensions.

\section{The derivation of upper bound}

In order to make the CNFs model trainable, we consider optimizing the upper bound of $\mathcal{L}_\textup{NLL}$. Consider the term
$\textup{tr} (\frac{\partial f(z(t),t;\theta)}{\partial z(t)}),$ by applying Theorem 1, we have

\begin{equation}
    \textup{tr} (\frac{\partial f(z(t),t;\theta)}{\partial z(t)})  \leq \sqrt{n} \left\| \frac{\partial f(z(t),t;\theta)}{\partial z(t)} \right\|_F,
\end{equation} where $n$ is the dimension of the Jacobian matrix of $ f(z(t),t;\theta)$ on $z(t)$.
By applying Theorem 2, we have

\begin{equation}
    \left\| \frac{\partial f(z(t),t;\theta)}{\partial z(t)} \right\|_F \leq \sqrt{n} \left\|\frac{\partial f(z(t),t;\theta)}{\partial z(t)} \right\|.
\end{equation} 

Hence,

\begin{equation}
    \textup{tr} (\frac{\partial f(z(t),t;\theta)}{\partial z(t)})  \leq \sqrt{n} \left\| \frac{\partial f(z(t),t;\theta)}{\partial z(t)} \right\|_F \leq n \left\|\frac{\partial f(z(t),t;\theta)}{\partial z(t)} \right\|.
\end{equation}

Previously, we assumed that $f(z(t),t;\theta)$ is uniformly Lipschitz continuous in $z(t)$. Also the function $z(t)$ is continuous and everywhere differentiable on $t$. From Theorem 4, for $\forall f(z(t),t;\theta)$ uniformly Lipschitz continuous in $z(t)$, $\exists L_{\theta} \in \mathbb{R} $ and $ L_{\theta} \geq 0$:

\begin{equation}
    \left\| \frac{\partial f(z(t),t;\theta)}{\partial z(t)} \right\| \leq L_{\theta},
\end{equation} where $\theta$ is the parameter of $f$.
Also, we have

\begin{equation}
    \textup{tr} (\frac{\partial f(z(t),t;\theta)}{\partial z(t)}) \leq n \left\|\frac{\partial f(z(t),t;\theta)}{\partial z(t)} \right\|,
\end{equation}

hence,

\begin{equation}
    \int_0^1 \textup{tr} (\frac{\partial f(z(t),t;\theta)}{\partial z(t)}) dt \leq n \int_0^1 \left\| \frac{\partial f(z(t),t;\theta)}{\partial z(t)} \right\| dt \leq nL_{\theta}.
\end{equation}

Also recall $f$ is uniformly Lipschitz continuous, hence there exists a minimal Lipschitz constant:

\begin{equation}
    \widehat{L_{\theta}}=\sup\limits_{z(t)} \left\| \frac{\partial f(z(t),t;\theta)}{ \partial z(t)} \right\|.
\end{equation}

Therefore $\mathcal{L}_\textup{NLL}$ exists an upper bound:

\begin{equation}
    \mathcal{L}_{\text{UB}} = -\log p(z(0))+ n \sup\limits_{z(t)} \left\| \frac{\partial f(z(t),t;\theta)}{ \partial z(t)} \right\|.
\end{equation}

It is not difficult to find that our optimization goal is mainly determined by $\widehat{L_{\theta}}$. Also, the spectral norm of all matrix is always non-negative. Therefore, when $\sup\limits_{z(t)} \left\| \frac{\partial f(z(t),t;\theta)}{ \partial z(t)} \right\|=0$, the upper bound $\mathcal{L}_{\text{UB}}$ reaches its minimum value.

%
% \begin{thebibliography}{8}
% \bibitem{ref_article1}
% Author, F.: Article title. Journal \textbf{2}(5), 99--110 (2016)

% \bibitem{ref_lncs1}
% Author, F., Author, S.: Title of a proceedings paper. In: Editor,
% F., Editor, S. (eds.) CONFERENCE 2016, LNCS, vol. 9999, pp. 1--13.
% Springer, Heidelberg (2016). \doi{10.10007/1234567890}

% \bibitem{ref_book1}
% Author, F., Author, S., Author, T.: Book title. 2nd edn. Publisher,
% Location (1999)

% \bibitem{ref_proc1}
% Author, A.-B.: Contribution title. In: 9th International Proceedings
% on Proceedings, pp. 1--2. Publisher, Location (2010)

% \bibitem{ref_url1}
% LNCS Homepage, \url{http://www.springer.com/lncs}, last accessed 2023/10/25
% \end{thebibliography}
\end{document}